\def\year{2021}\relax

\documentclass[letterpaper]{article}
\usepackage{aaai21}
\usepackage{times}
\usepackage{helvet} 
\usepackage{courier}  
\usepackage[hyphens]{url}  
\usepackage{graphicx} 
\usepackage{natbib}
\usepackage{caption}
\usepackage{amsfonts}
\usepackage{booktabs}
\usepackage{multirow}
\usepackage{amsmath}
\usepackage{hhline}
\usepackage{colortbl}
\usepackage{pgf-pie}
\usepackage{pgfplots}
\pgfplotsset{compat=1.13}
\usepackage{enumerate}[nosep]
\usepackage[normalem]{ulem}
\usepackage{hyperref}

\usepackage[switch]{lineno}

\newcommand{\hlc}[2][green]{{\sethlcolor{#1}\hl{#2}}}

\newcommand{\prem}{$\mathcal{P}$}
\newcommand{\hyp}{$\mathcal{H}$}
\newcommand{\upd}{$\mathcal{U}$}
\newcommand{\ut}{$\mathcal{T}$}
\newcommand{\rat}{$\mathcal{R}$}
\newcommand{\dataset}{e-$\delta$-NLI}
\newcommand{\task}{$\delta$-NLI}
\newcommand{\mprem}{\mathcal{P}}
\newcommand{\mhyp}{\mathcal{H}}
\newcommand{\mupd}{\mathcal{U}}

\newcommand{\mrat}{\mathcal{R}}

\usepackage{color}
\usepackage{xcolor}
\usepackage{soul}
\definecolor{orange}{RGB}{255, 50, 190}
\definecolor{green}{RGB}{228, 249, 189}
\definecolor{deeppeach}{rgb}{1.0, 0.8, 0.64}

\title{
Learning to Rationalize for Nonmonotonic Reasoning with Distant Supervision
}

\author{Faeze Brahman$^1$\thanks{Work done during internship at AI2.}, Vered Shwartz$^{2,3}$, Rachel Rudinger$^4$, and Yejin Choi$^{2,3}$\\}

\affiliations{
$^1$University of California, Santa Cruz\\
$^2$Allen Institute for AI\\
$^3$Paul G. Allen School of Computer Science \& Engineering, University of Washington\\  $^4$University of Maryland, College Park, MD\\

{\tt fbrahman@ucsc.edu, rudinger@umd.edu, \{vereds,yejinc\}@allenai.org} \\
}

\begin{document}


\maketitle

\begin{abstract}

The black-box nature of neural models has motivated a line of research that aims to generate natural language rationales to explain why a model made certain predictions. Such rationale generation models, to date, have been trained on dataset-specific crowdsourced rationales, but this approach is costly and is not generalizable to new tasks and domains. In this paper, we investigate the extent to which neural models can reason about natural language rationales that explain model predictions, relying only on distant supervision with no additional annotation cost for human-written rationales. We investigate multiple ways to automatically generate rationales using pre-trained language models, neural knowledge models, and distant supervision from related tasks, and train generative models capable of composing explanatory rationales for unseen instances. We demonstrate our approach on the defeasible inference task, a nonmonotonic reasoning task in which an inference may be strengthened or weakened when new information (an update) is introduced. Our model shows promises at generating post-hoc rationales explaining why an inference is more or less likely given the additional information, however, it mostly generates trivial rationales reflecting the fundamental limitations of neural language models. Conversely, the more realistic setup of jointly predicting the update or its type and generating rationale is more challenging, suggesting an important future direction.

\end{abstract}

\section{Introduction}
\label{sec:intro}

Deep neural models perform increasingly well across NLP tasks, but due to their black-box nature, their success comes at the cost of our understanding of the system. The lack of transparency for \emph{why} a model made a particular prediction may---among other problems---introduce fairness issues \cite{fairness}, and hide the fact that often a model is right for the wrong reasons due to learning dataset-specific shortcuts and annotation artifacts \cite{gururangan-etal-2018-annotation, poliak-etal-2018-hypothesis}. 
There is growing interest in NLP in opening the black-box, through surrogate models \cite{ribeiro2016should}, counterfactual evaluation \cite{tenney-etal-2020-language}, examining the inner structure of the neural network \cite{Raffel17, JainWPW20}, or generating natural language explanations. We focus on the latter approach. Recent work by \citet{NIPS2018_8163} and \citet{rajani18} collected human-written explanations for the natural language inference \cite[NLI;][]{bowman2015large} and commonsense question answering \cite[CommonSenseQA;][]{talmor-etal-2019-commonsenseqa} tasks and trained models to predict explanations for new instances. Such supervision is not always accessible, is expensive to obtain, and is unlikely to generalize well across datasets. 



\begin{figure}[t]
\centering
\includegraphics[width=.9\linewidth]{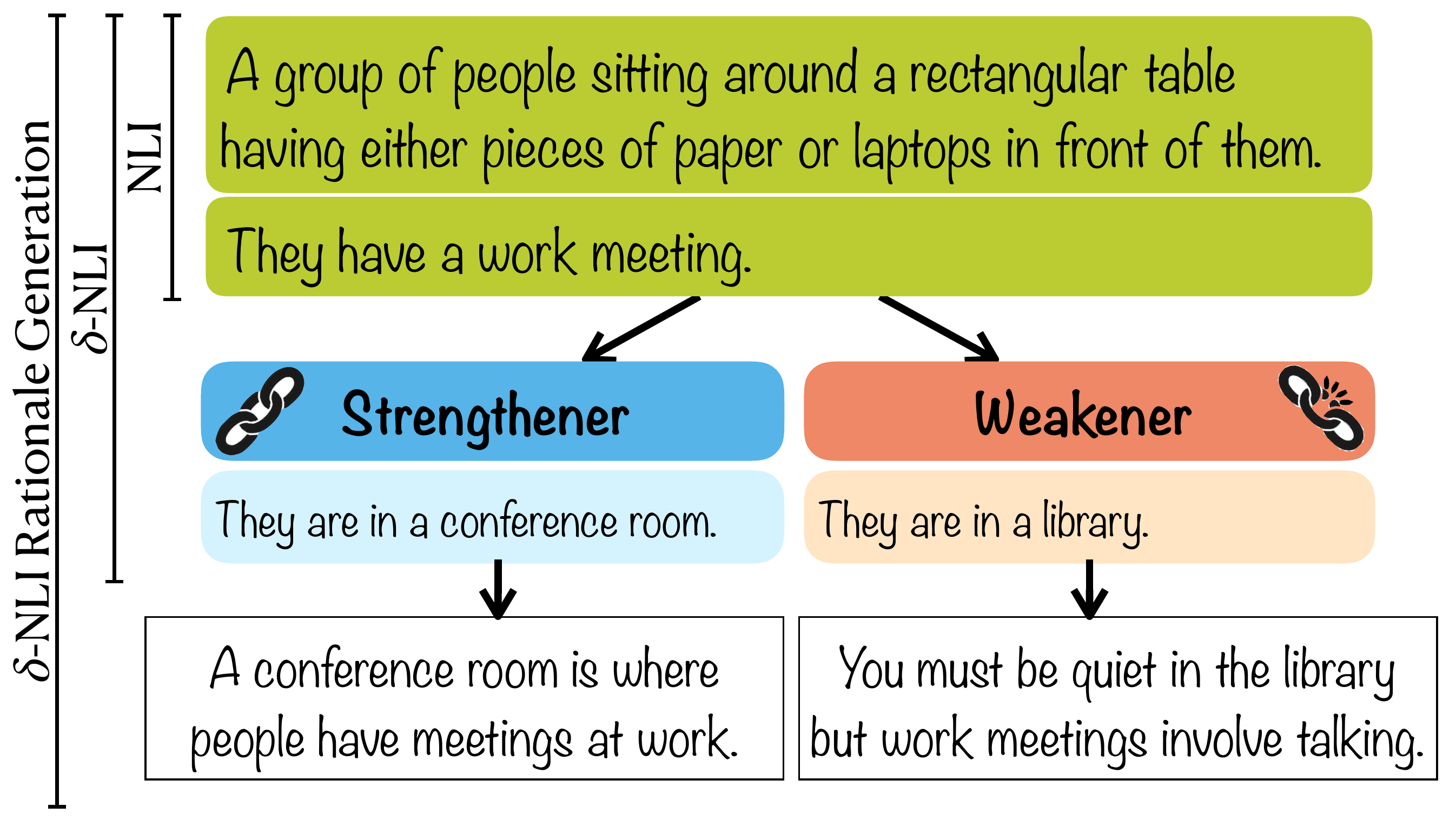}
\vspace{-10pt}
\caption{An illustration of the NLI, \task{} (Defeasible NLI), and \task{} Rationale Generation tasks.}
\label{fig:task}
\end{figure}
 
In this work, we explore learning to rationalize using a distant supervision approach \emph{without additional annotation cost}. We focus on the Defeasible Inference task \cite[\task{}; ][]{definf2020}, illustrated in Figure~\ref{fig:task}. Given premise and hypothesis sentences, and an update sentence, the goal of the (discriminative) \task{} task is to recognize whether the update weakens or strengthens the entailment of the hypothesis by the premise. For example, the update that ``the people are in a conference room'' strengthens the hypothesis that ``they are in a work meeting''. An alternative (generative) task is to generate the update given the premise, hypothesis, and update type (strengthener or weakener).  

\begin{figure*}[t]
\centering
\includegraphics[width=.9\linewidth]{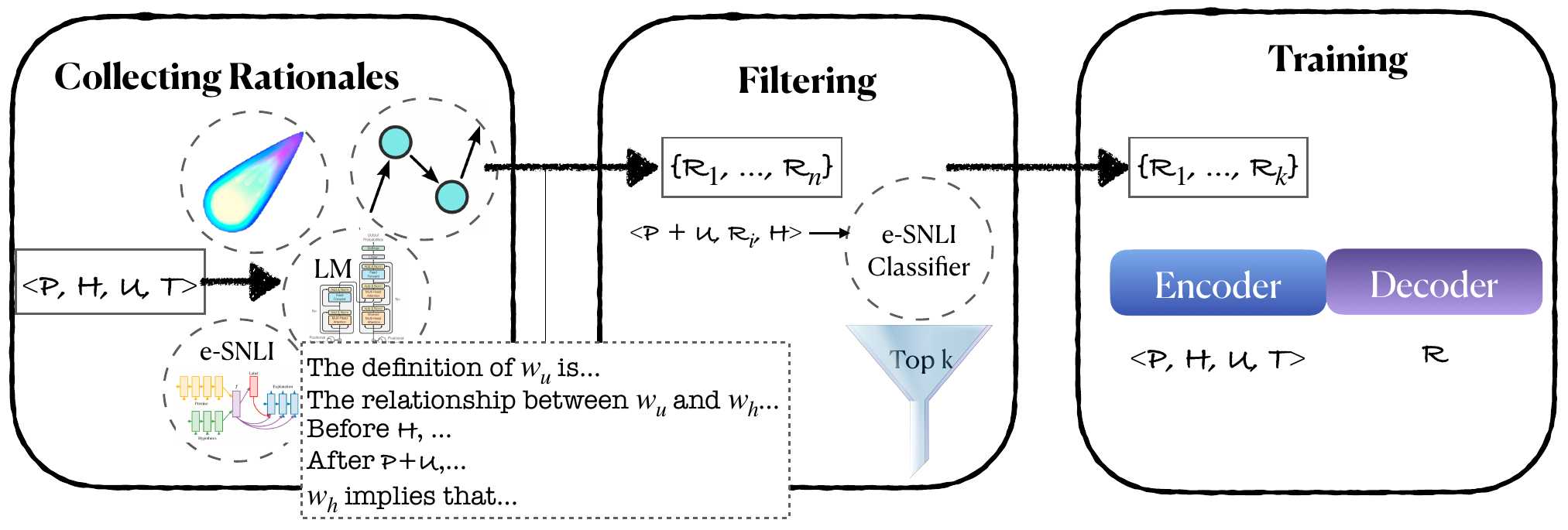}
\vspace{-5pt}
\caption{The complete training process: (1) collecting rationales from various sources, (2) Keeping the top k most helpful rationales; (3) training a generative model. During inference, we apply the generative model directly to the inputs.}
\label{fig:training}
\end{figure*}

We present the Defeasible Inference Rationale Generation task, with the goal of generating natural language rationales that explain why a hypothesis is \emph{more} likely after learning about a \emph{strengthener} update and \emph{less} likely after learning about a \emph{weakener} update. To that end, we create the \dataset{} dataset by augmenting the \task{} dataset with rationales from various sources, including pre-trained language models, knowledge bases, and supervision from a related task. We then train two types of language model-based rationale generation models: post-hoc models that generate a rationale given access the target values (i.e., the update or update type); and joint models that jointly generate the target value along with the rationale. The overall workflow of our approach is shown in Figure~\ref{fig:training}.


We evaluate the models with both automatic and human evaluations. The results of the post-hoc models are promising, with most generated rationales considered relevant and factually correct, and 40\% on average considered explanatory. In line with prior work by \citet{NILE20}, further analysis revealed that models trained to post-hoc rationalize develop strategies to trivially map the target value to one of the several patterns associated with it in the training data, such as ``the update implies that hypothesis''.

We consider the joint setup, in which the model has no access to the target value, to be more realistic. On this challenging setup, that hinders the models' ability to learn trivial shortcuts, the performance is worse, warranting future research in this direction.\footnote{The code and data are available at: \url{https://github.com/fabrahman/RationaleGen}.}

\section{Background}
\label{sec:background}

\paragraph{Natural Language Inference.} Recognizing Textual Entailment \cite[RTE;][]{2013Dagan}, or, in its newer variant, Natural Language Inference \cite[NLI;][]{bowman2015large}, is defined as a 3-way classification task. Given a premise sentence \prem{} and a hypothesis sentence \hyp{}, the goal is to determine whether \prem{} entails, contradicts, or is neutral with \hyp. \prem{} is said to entail \hyp{} if a human reading \prem{} would typically infer that \hyp{} is most likely true. \hyp{} is neutral if it could be but is not necessarily true given \prem. 

In recent years, several large-scale datasets for the task have been released  \cite[e.g.][]{williams2018broad,nie2019adversarial}, encouraging training neural models. We focus on the Stanford Natural Language Inference dataset \cite[SNLI;][]{bowman2015large}, in which image captions serve as premises, and hypotheses were crowdsourced. 



\paragraph{Explainable NLI.} Since deep learning has become the dominant paradigm in NLP research, efforts have been devoted to opening the ``black-box'' and interpreting neural models' predictions. One approach looks into the model's weights and traces back salient spans from the input that affected the prediction. The attention mechanism \cite{bahdanau:14}, which is popular across NLP models, facilitates this through the attention weights \cite{Raffel17, JainWPW20}. However, whether or not attention weights provide reliable insights into the model's decision-making process is still debatable \cite{serrano-smith-2019-attention, jain-wallace-2019-attention, wiegreffe-pinter-2019-attention}.

An alternative approach is to generate natural language explanations for the model's decisions. This is typically done by training a model on free-form human explanations \cite{NIPS2018_8163, rajani18, wang-etal-2019-make, zellers2019neuralfakenews}, however, such supervision is not always available, and is costly to obtain. To that end, we propose a distant supervision approach that requires no additional supervision. Among other data source, we leverage the e-SNLI dataset \cite{NIPS2018_8163}, in which premise-hypothesis pairs from SNLI have been augmented with human-written explanations for the gold labels.

There are several setups for interpretation methods: (i) \textbf{ante-hoc}: generating the rationale from the input, and providing it to the decision-making model with the input \cite{lei-etal-2016-rationalizing, bastings-etal-2019-interpretable, NILE20} or without it \cite{JainWPW20}; (ii) \textbf{joint}: generating the rationale and the label jointly \cite{wt5}; and (iii) \textbf{post-hoc}: generating a rationale given the input and the gold or predicted label. The motivation for the first approach is to produce faithful rationales, i.e. rationales representing the model's true decision process. However, there is no guarantee that the decision-making model actually uses the rationales. Moreover, in some cases the selected rationale is not sufficient to make the prediction without the input \cite{wiegreffe2020measuring}, while in others, label-specific rationale templates may make the label prediction trivial given the rationale \cite{NILE20}. We focus on the latter two approaches: joint and post-hoc, while acknowledging that our rationales are not constructed to be faithful.\footnote{Humans also post-hoc rationalize decisions, and it is known to be flawed \cite{gazzaniga2013integrated}. For recent works discussing rationale faithfulness, see \citet{hase-bansal-2020-evaluating}, \citet{jacovi-goldberg-2020-towards}, and \citet{wiegreffe2020measuring}.}

\paragraph{Defeasible Inference.} Defeasible reasoning is a nonmonotonic logic in which valid inferences can become invalid when new information is introduced. For example, ``Tweety is a bird'' entails that ``Tweety flies'' unless provided with additional information such as ``Tweety is a penguin'' \cite{reiter1980logic}. Despite being a fundamental mode of human reasoning, modern NLP research paid little attention to nonmonotonic reasoning \cite[e.g.][]{qin2019counterfactual,bhagavatula2019abductive}. Recently, \citet{definf2020} coupled defeasible reasoning with natural language inference by adding an update sentence \upd{} to the premise \prem{} and hypothesis \hyp. Expanding the traditional definition, \upd{} may either \emph{weaken} or \emph{strengthen} \hyp. 


Two defeasible inference (\task{}) tasks were introduced: \emph{discriminative defeasible inference}, in which given \prem, \hyp, and \upd, the goal is to classify the update as either weakener or strengthener (update type, \ut); and \emph{generative defeasible inference}, in which given \prem, \hyp, and \ut{} the goal is to generate an update \upd{} with the required type. The dataset for these tasks was built by crowdsourcing update sentences for neutral sentence-pairs from existing NLI datasets, Specifically, here we use the SNLI portion of their data. 


\paragraph{Unsupervised Knowledge Extraction from Pre-trained LMs.} Pre-trained Language Models (LMs) based on the neural transformer architecture \cite{Vaswani:17}, such as GPT2 \cite{radford2019language} and BERT \cite{devlin-etal-2019-bert} have greatly improved the performance on NLP tasks that require world knowledge and commonsense reasoning. While the best practice is to fine-tune the LM, they may also be used in an unsupervised manner. \citet{petroni-etal-2019-language} and \citet{davison-etal-2019-commonsense} completed commonsense knowledge bases (KB) by converting triplets into free-form text and predicting or scoring the target concept. \citet{tamborrino-etal-2020-pre} leveraged masked LMs to score the plausibility of answer choices in multiple-choice commonsense question answering (QA) tasks. \citet{shwartz2020unsupervised} used LMs to generate information-seeking clarification questions (e.g. ``What is the definition of...'') and their answers for providing relevant knowledge for commonsense QA tasks, which yielded similar performance gains to models utilizing KBs. Similarly, \citet{latcinnik2020explaining} used LMs to generate a textual hypothesis which was used by the answer scorer of a multiple choice QA task.

\section{\dataset{} Dataset}
\label{sec:data}
We now describe \dataset{} (Explanations for Defeasible NLI). We augmented the \task{} dataset described in Section \ref{sec:background} with rationales that explain why a hypothesis is \emph{more} likely after learning about a \emph{strengthener} update and \emph{less} likely after learning about a \emph{weakener} update. Rather than eliciting rationales from humans, we take a distant supervision approach and gather rationales from various sources, as exemplified in Table~\ref{tab:rationale_examples} and described below. 

\subsection{Collecting Rationales}
\label{sec:data:distant_supervision}

Certain spans in the inputs \hyp{} and \upd{} are highly salient for classifying the update type in the discriminative \task{} task. We hypothesize that these same spans will be salient for the task of generating rationales. Therefore we use the \task{} update type classifier and score each token in the input by its attention weight from the \texttt{<cls>} token in the final layer, and extract the set of top 20\% non-continuous spans with respect to that score, denoted as $S$. For example, in Table~\ref{tab:rationale_examples}, the most salient spans are highlighted in orange (hypothesis) and green (update). We use the following sources to extract or generate rationales. 


\begin{table*}[t]
    \scriptsize
    \centering
    \setlength\tabcolsep{1.5pt}
\begin{tabular}{p{1.35cm}p{8cm}p{7.5cm}}
\toprule
\textbf{Source} & \textbf{Instance} &  \textbf{Rationales} \\ 
\toprule
\multirow{3}{*}{\parbox{1.35cm}{\textbf{Vanilla LM}}} & $\mathcal{P}$: [...] pedestrians walking down street filled with vendors and umbrella carts. & \multirow{3}{*}{\parbox{7.5cm}{The relationship between ``a busy Manhattan sidewalk selling hotdogs'' and ``weekly farmer's market'' is that they both exist in tandem, but not necessarily together.}}\\ 
& $\mathcal{H}$: The vendors are there for \hlc[deeppeach]{the} weekly \hlc[deeppeach]{farmer's market.} & \\
& \textbf{W}: They are \hlc[green]{on} a busy \hlc[green]{Manhattan} sidewalk selling \hlc[green]{hotdogs.} \\ 
\midrule
\multirow{3}{*}{\parbox{1.35cm}{\textbf{KG-Enhanced LM}}} & $\mathcal{P}$: A person wearing red and white climbs a foggy mountain. & The purpose of ``rock climbing'' is to reach a high place. \\
& $\mathcal{H}$: \hlc[deeppeach]{A} person \hlc[deeppeach]{is} rock \hlc[deeppeach]{climbing.} & \multirow{2}{*}{\parbox{7.5cm}{The relationship between ``rope'' and ``climbing'' is that rope has property used to climb.}} \\ 
& \textbf{S}: The person is attached to a \hlc[green]{rope going up} the side of the \hlc[green]{mountain.} &  \\ 
\midrule
\multirow{3}{*}{\parbox{1.35cm}{\textbf{COMeT}}} & $\mathcal{P}$: A baby boy in an elmo chair with lots of toys in the background. & $\mathcal{H}$ precondition: The baby boy is seen as joyful. \\
& $\mathcal{H}$: The baby boy in the elmo chair is happy. & $\mathcal{U}$ postconditions: As a result, boy's mom feels to console. \\
& \textbf{W}: The baby boy's mom is wiping tears from his eyes. \\
\midrule
\multirow{3}{*}{\parbox{1.35cm}{\textbf{NLI-derived}}} & $\mathcal{P}$: The brown dog catches a ball in the air. & Catching a ball in the air implies that the dog plays with the ball. \\
& $\mathcal{H}$: The dog plays with the ball outside. & Bushes are outside. \\
& \textbf{S}: The ball skips into the bushes. \\ 
\midrule
\multirow{3}{*}{\parbox{1.35cm}{\textbf{NLI-derived w/ \\Highlights}}} & $\mathcal{P}$: A woman wearing [...] and sunglasses, walks through a shopping outlet. & If a woman is carrying bags, then she is buying goods. \\
& $\mathcal{H}$: The \hlc[deeppeach]{woman} is \hlc[deeppeach]{buying goods.} \\
& \textbf{S}: The woman \hlc[green]{is carrying} shopping \hlc[green]{bags.}\\ 
\bottomrule
\end{tabular}
\setlength\tabcolsep{6pt}

    \vspace{-5pt}
    \caption{Examples of rationales generated from each of the sources. \textbf{W} stands for a weakener update and \textbf{S} for strengthener.}
    \label{tab:rationale_examples}
\end{table*}

\paragraph{Vanilla LM.} We generate two types of rationales: definitions and purposes for single spans, and relationships for a pair of spans. We use SpaCy \cite{honnibal2017spacy} to keep only the grammatical salient spans $S_G \subseteq S$ by filtering out stop words and keeping both the entire (noun or verb) phrase and its head for each span.

Following \citet{shwartz2020unsupervised}, we prompt the LM with ``[context]. The definition of $np$ is'' for each noun phrase in $S_G$, and ``[context]. The purpose of $vp$ is'' for verb phrases in $S_G$. We set the context to the concatenation of premise and hypothesis (\prem + \hyp{}) when the target phrase is in the hypothesis, and to \prem + \upd{} when it is from the update. 

In addition, we generate the relationship between pairs of spans. We take the top 3 most similar pairs of $s_u$ (subset of $S_G$ originated from \upd{}) and $s_h$ (subset of $S_G$ originated from \hyp{}), judged by the cosine similarity between their word2vec embeddings \cite{mikolov2013efficient}.\footnote{For multi-word spans we use maximum word-level similarity.} We prompt the LM with ``\prem + \upd + \hyp{}. The relationship between s$_u$ and s$_h$ is that''. 

We use GPT2-M \cite{radford2019language} via the Transformers package \cite{Wolf2019HuggingFacesTS}. We limit the rationale length to up to 12 tokens, and use Nucleus sampling \cite{Holtzman2020The} with $p = 0.35$, and $\text{temperature} = 1.0$ to generate at most 20 rationales for each prompt.\footnote{Hyper-parameter values were chosen empirically from $p \in \{0.35,0.5,0.75\}$, $\text{temperature} \in \{0.7,1\}$, $\#\text{samples} \in \{5,20\}$.}

\paragraph{Knowledge-Enhanced LM.} To further instill commonsense knowledge into the LM, we follow \citet{guan2020knowledge} and continue pre-training GPT2-M on triplets from ConceptNet \cite{speer2017conceptnet} converted to natural language using the templates from \citet{davison-etal-2019-commonsense}. For example, \texttt{(a glass of milk, UsedFor, drinking)} is converted to ``A glass of milk is used for drinking''. We train the LM on the transformed triplets for 2 epochs. We then use the LM as previously detailed to generate definitions, purposes, and relationships. We use Nucleus sampling with $p = 0.5$, $\text{temperature} = 0.7$, and generate up to 5 rationales for each prompt.

\paragraph{COMeT.} COMeT \cite{bosselut-etal-2019-COMeT} is a LM-based knowledge base completion model. We use the model trained on ATOMIC \cite{sap2019atomic}, a commonsense KB consisting of \textit{if-then} triplets concerning everyday situations, along multiple dimensions. We generate the postconditions following the update (\texttt{xWant}, \texttt{xEffect}, \texttt{xReact}, \texttt{xAttr}, \texttt{oWant}, \texttt{oEffect}, \texttt{oReact}) and the preconditions that lead to the 
hypothesis (\texttt{xNeed}, \texttt{xIntent}, \texttt{xAttr}). We use beam search with beam size of 5 as the decoding strategy, keeping the entire beam, and replace \texttt{PersonX} with the syntactic subject of the input sentence.



\begin{figure}[t]
    \centering
    \definecolor{one}{HTML}{f0f9e8}
\definecolor{two}{HTML}{bae4bc}
\definecolor{three}{HTML}{7bccc4}
\definecolor{four}{HTML}{43a2ca}
\definecolor{five}{HTML}{0868ac}

\begin{tikzpicture}[scale=0.8]
\scriptsize
\def\printonlypositive#1{\ifdim#1pt>5pt
#1
\fi}
\tikzset{
     lines/.style={draw=none},
}
\pie[
    /tikz/every pin/.style={align=left},
    text=legend,
    radius=1.7, 
    style={lines}, 
    after number=,
    color = {one,two,three,four,five}]
{
41.05/Knowledge-Enhanced LM, 
37.91/COMeT,
9.16/Vanilla LM,
7.1/NLI-derived,
4.77/NLI-derived w\/ highlights
}
\end{tikzpicture} \\~\\
    \begin{tikzpicture}[scale=0.8]
\begin{axis}[
    xbar stacked,
    ytick=data,
    axis y line*=none,
    axis x line*=bottom,
    tick label style={font=\footnotesize},
    label style={font=\tiny},
    xtick={},
    width=.5\textwidth,
    bar width=4mm,
    yticklabels={},
    xmin=0,
    xmax=100,
    ymin=-0.6,
    xtick style={draw=none},
    area legend,
    y=6mm,
    nodes near coords=\pgfmathprintnumber{\pgfplotspointmeta}\%,
    nodes near coords style={text=black, at ={(\pgfplotspointmeta,\pgfplotspointy)},anchor=west},
    visualization depends on=y \as \pgfplotspointy,
    every axis plot/.append style={fill}
]
\addplot[five] coordinates
  {(39.07,0) (0,1) (0,2) (0,3) (0,4)};
\addplot[four] coordinates
  {(0,0) (57.94,1) (0,2) (0,3) (0,4)};
\addplot[three] coordinates
  {(0,0) (0,1) (22.76,2) (0,3) (0,4)};
\addplot[two] coordinates
  {(0,0) (0,1) (0,2) (17.12,3) (0,4)};
\addplot[one] coordinates
  {(0,0) (0,1) (0,2) (0,3) (12.16,4)};
\end{axis}  
\end{tikzpicture}
    \vspace{-5pt}
    \caption{\textbf{Top:} Percentage of each source among the rationales in the final \dataset{} dataset. \textbf{Bottom:} Percentage of rationales that remained after filtering from each source.}
    \label{fig:sources}
\end{figure}
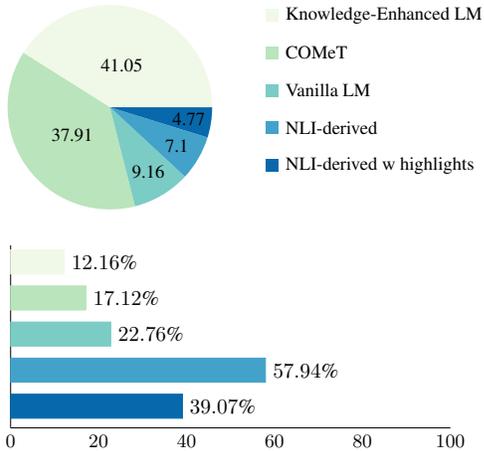

\paragraph{NLI-derived.} We repurpose a model for the related task of NLI rationale generation for our task of rationale generation for \task{}. To that end, we reproduced the WT5 model suggested by \citet{wt5}. The model is based on the T5 encoder-decoder language model \cite{2019t5}, and is trained on the e-SNLI dataset \cite{NIPS2018_8163} to jointly generate the label (entailment, contradiction) and the rationale for a given premise and hypothesis pair. More concretely, the input consists of the task prefix and the inputs (\texttt{explain nli premise: \prem{} hypothesis: \hyp}) while the expected output is \texttt{label explanation: \rat{}}. During inference, we set the premise to \prem + \upd, i.e., treating the update as part of the premise, and provide it to the model along with \hyp. The model generates the binary entailment label (excluding neutral) between \prem + \upd{} and \hyp, and the rationale that explains the label.\footnote{In practice, we only take the rationales and ignore the labels. But if we map entailment to strengthener and contradiction to weakener, we get 64\% accuracy on the update type prediction. We note that this is an approximation. The definition of defeasible inference requires that a weakener makes the hypothesis \emph{less} likely but not necessarily \emph{unlikely}, while a strengthener makes the hypothesis \emph{more} likely but not necessarily \emph{likely}.} 

\paragraph{NLI-derived with Highlights.} Each instance in e-SNLI highlights salient spans in the input that the annotators considered helpful for explaining the label. We train a variant of the T5-based e-SNLI model that gets (only) the highlighted words as input and outputs the label and the rationale. We then generate rationales for the \task{} dataset by applying the model to salient spans in $S_G$ that originated in \upd{} or \hyp{}.

\subsection{Filtering Rationales}
\label{sec:data:filtering}

Following the collection step, each instance in the \task{} dataset is now augmented with a list of candidate rationales explaining its label (update type). To further improve the quality of this distant supervision, we rank and keep the best rationales. In particular, we would like to keep the rationales that are most helpful for predicting the label. Ideally, we would want to train a \task{} classifier that gets \prem, \hyp, \upd, and the rationale as input and outputs the update type. However, this causes a circular problem because we don't yet know which rationales are reliable. 

\begin{table*}[t]
    \scriptsize
    \centering
    \setlength\tabcolsep{4pt}
    \begin{tabular}{llll}
    \toprule
    \textbf{Task} & \textbf{Objective} & \textbf{Input} & \textbf{Output} \\ 
    \midrule
    \rowcolor{blue!10} \multicolumn{4}{c}{\textbf{Post-hoc Rationalization}} \\
    (1) Rationale & $P$(\rat $|$ \prem, \hyp, \ut, \upd) & \texttt{[premise] \prem{} [hypo] \hyp{} [ut] $<$\ut$>$ [update] \upd{} [rationale]} & \rat \\ 
    (2) Update type & $P$(\ut $|$ \prem, \hyp, \upd, \rat) & \texttt{[premise] \prem{} [hypo] \hyp{} [update] \upd{} [rationale] \rat{} [ut]} & $<$\ut$>$ \\ 
    (3) Update & $P$(\upd $|$ \prem, \hyp,\ut, \rat) & \texttt{[premise] \prem{} [hypo] \hyp{} [rationale] \rat{} [ut] $<$\ut$>$ [update]} & \upd{}  \\
    (4) Multi & (1) + (2) + (3) &  \\
    \rowcolor{blue!10} \multicolumn{4}{c}{\textbf{Joint Prediction and Rationalization}} \\
    (5) Update type + Rationale & $P$(\ut, \rat $|$ \prem, \hyp, \upd) & \texttt{[premise] \prem{} [hypo] \hyp{} [update] \upd{} [update\_type\_rationale]} & \texttt{[ut] $<$\ut$>$ [rationale] \rat} \\ 
    (6) Update + Rationale & $P$(\upd, \rat $|$ \prem, \hyp, \ut) & 
    \texttt{[premise] \prem{} [hypo] \hyp{} [ut] $<$\ut$>$ [update\_rationale]} & \texttt{[update] \upd{} [rationale] \rat} \\ 
    \bottomrule
    \end{tabular}
\setlength\tabcolsep{6pt}
    \vspace{-5pt}
    \caption{The different training setups we experiment with. We add special tokens to mark the boundaries of each input and output span, e.g. \texttt{[premise]} marks the beginning of the premise.}
    \label{tab:train_setups}
\end{table*}

Hence, again we use e-SNLI as a proxy. We train a classifier on e-SNLI that gets the premise, hypothesis, and rationale as inputs and predicts the entailment label (entailment, contradiction). Specifically, we fine-tune a binary RoBERTa classifier \cite{liu2019roberta} with the following input format: \texttt{\prem{} \texttt{<sep>} \rat{} \texttt{<sep>} \hyp}. For a \task{} instance (\prem, \hyp, \ut, \upd) with a set of candidate rationales $\{ \mathcal{R}_i \}_{i=1}^{N_R}$ (of various sources), we compute: $o = \operatorname{NLI}(\mprem+\mupd$ \texttt{<sep>} $\mrat_i$\texttt{<sep>} $\mhyp)$, where $o$ is a 2-dimensional vector representing the confidence of the classifier in each label. We score each rationale by the confidence assigned to the label associated with its update type: strengtheners as entailment and weakener as contradiction\footnote{Equating strengtheners with entailment and weakeners as contradiction is a simplifying assumption, which is not strictly true.}, and rank the rationales accordingly. We keep the top 10\% ranked rationales for each instance, yielding 8 rationales per instance on average.

We follow the original split for train (80\%), test (10\%), and development (10\%) sets. By augmenting the data with multiple rationales per original \task{} instance, the final \dataset{} dataset consists of 731,579 training, 15,781 test, and 15,527 development instances. Figure~\ref{fig:sources} shows the percent of rationale sources in the dataset and rationales that remained after filtering from each source.

\section{Rationale Generation Model}
\label{sec:model}
We use the \dataset{} dataset to train various generative models with the goal of generating rationales that explain why a hypothesis is \emph{more} likely after learning about a \emph{strengthener} update and \emph{less} likely after learning about a \emph{weakener}. 

Every instance in the \dataset{} dataset consists of a premise \prem, hypothesis \hyp, update type \ut, update \upd, and a set of rationales $\{ \mathcal{R}_i \}_{i=1}^{N_R}$. During training, we treat every (\prem, \hyp, \ut, \upd, \rat) for \rat $\in \{ \mathcal{R}_i \}_{i=1}^{N_R}$ as a separate instance.

\subsection{Architecture and Implementation Details}
\label{sec:model:architecture}

We fine-tune transformer-based pre-trained LMs on the \dataset{} dataset. Specifically, we use GPT2-XL \cite{radford2019language} and Bart-L \cite{lewis2019bart}.\footnote{In our preliminary experiments, we also experimented with T5, but we did not observe any improvements.} We use the Transformers package \cite{Wolf2019HuggingFacesTS}, training each model for two epochs with batch size of 8 (GPT2), and 128 (Bart) on a Quadro RTX 8000 GPU machine.


\subsection{Training Objective}
\label{sec:model:objective}


We minimize the conditional log-likelihood of the output given the input: $\mathcal{L} = -\sum_{i=1}^{n} \log p(x^{out}_i|{x^{out}_{<i}, x^{in}})$. In particular, for GPT2, which is a standard LM model, the loss is computed over the entire sequence $[x^{in};x^{out}]$, whereas in Bart, which is an encoder-decoder model, the loss is computed only over the output sequence, $x_{out}$. 

We experiment with various training setups described in Table~\ref{tab:train_setups}. Our setups can be divided into two categories. The first category is \textbf{Post-hoc Rationalization}, in which the model has access to the target values (i.e., update and update type) and is required to explain it. Our main task in this category is Rationale Generation (1). It is formulated as generating a rationale conditioned on the premise, hypothesis, update, and update type. Similarly, we can generate each of the update type (2) and update (3) given all other fields. These two setups are orthogonal to our goal, but we combine them with (1) in a multi-task setup (4) where we expect them to improve the model's generalizability \cite{shwartz-dagan-2018-paraphrase,zellers2019neuralfakenews} and improve the performance on the main task. The second and more realistic category is \textbf{Joint Prediction and Rationalization}, in which the model jointly predicts either update type (5) or update (6) along with an explanation. 


\section{Results}
\label{sec:results}
\begin{table}[t]
\scriptsize
    \centering

\setlength\tabcolsep{2pt}
    \begin{tabular}{llll p{0.0075cm} llll}
    \toprule
    \textbf{Objective} & \textbf{Model} & \multicolumn{2}{c}{\textbf{Automatic}} & & \multicolumn{4}{c}{\textbf{Human (\%)}} \\ 
    \hhline{~~--~----}
    & & \textbf{BLEU} & \textbf{ROUGE} & & \textbf{Gram.} & \textbf{Rele.} & \textbf{Corr.} & \textbf{Expl.} \\ 
    \midrule
    \rowcolor{blue!10} \multicolumn{9}{c}{\textbf{Post-hoc Rationalization}} \\
    \multirow{2}{*}{\textbf{Rationale}} & GPT2-XL & 33.0 & 33.91 & & 92.5/93.5 & 33.5/60.0 & 52.5/45.5 & 2.0/4.5 \\ 
    & BART-L & 13.15 & 22.48 & &  \textbf{95.0/99.5} & \textbf{80.0/80.5} & \textbf{55.0/58.0} & \textbf{47.0/33.0} \\ 
    \midrule
    \multirow{2}{*}{\textbf{Multi}} & GPT2-XL & \textbf{33.58} & \textbf{34.50} & & 88.5/94.5 & 30.0/55.0 & 47.0/43.0 & 0.5/7.0 \\ 
    & BART-L & 17.38 & 24.03 & & 95/97.5 & 74.5/76.5 & 55.5/53.0 & 44.0/22.5 \\    
    \midrule
    \rowcolor{blue!10} \multicolumn{9}{c}{\textbf{Joint Prediction and Rationalization}} \\
    \textbf{Update} & GPT2-XL & 23.93 & \textbf{31.71} & & 85.0/88.0 & 15.5/35.0 & 30.5/15.5 & 1.0/1.5 \\ 
    \textbf{+ Rationale} & BART-L & 25.24 & 30.83 & & 86.5/83.0 & 20.0/34.5 & 34.0/17.5 & 2.5/0.5 \\    
    \midrule
    \textbf{Update type} & GPT2-XL & \textbf{27.90} & 31.18 & & 86.5/89.0 & \textbf{27.0/39.0} & \textbf{36.5/29.0} & \textbf{8.5/3.0} \\ 
    \textbf{+ Rationale} & BART-L & 24.54 & 29.04 & & 86.5/85.5 & 26.0/18.5 & 35.5/30.5 & 7.0/1.0 \\    
    \bottomrule
    \end{tabular}
\setlength\tabcolsep{6pt}

    \vspace{-5pt}
    \caption{Automatic and human evaluation of rationale generation for the test set. Human evaluation results are presented for strengtheners and weakeners separately (S/W).}
    \label{tab:results}
\end{table}

For each combination of rationale generation training setup, we generated a rationale for each instance in the test set using beam search with 5 beams. We evaluated the generated rationales both in terms of automatic metrics and human evaluation. The results are shown in Table~\ref{tab:results}. 


\subsection{Automatic Evaluation}
\label{sec:results:auto}

We used standard n-gram overlap metrics: the precision-oriented BLEU score \cite{papineni2002bleu} and recall-oriented ROUGE score \cite{lin2004rouge}. Specifically, we used BLEU-4 that measures overlap of n-grams up to $n=4$, and ROUGE-L that measures longest matching sequences, and compared multiple predictions against multiple distantly supervised rationales as references. The result of the automatic measures are reported in Table~\ref{tab:results}. In general, GPT2-based models achieve better automatic scores. We also observe additive gain using multi-task setup on both BLEU and ROUGE scores.  

\subsection{Human Evaluation}
\label{sec:results:human}

Since automatic metrics have demonstrated low correlation with human judgments across various NLG tasks \cite{novikova2017we}, and because our automatic metrics only evaluate the generated rationales against the distantly supervised rationales (in place of human-written references), we also conduct a more reliable evaluation using human judges on Amazon Mechanical Turk. We sampled 200 instances, along with a generated rationale for each model. Following \citet{shwartz2020unsupervised}, we asked workers to determine whether a rationale was 1) grammatical, not entirely grammatical but understandable, or completely not understandable; 2) relevant to the instance (\prem, \hyp, and \upd); 3) factually correct or likely true; and 4) explanatory of the update type (i.e. why the strengthener makes the hypothesis more likely or the weakener makes it less likely). 
To ensure the quality of annotations, we required that the workers be located in the US, UK, or Canada, and have a 99\% approval rate for at least 5,000 prior tasks. We aggregated annotations from 3 workers using majority vote. The annotations yielded fair levels of agreement, with Fleiss' Kappa \cite{landis1977measurement} between $\kappa = 0.22$ for relevance and $\kappa = 0.37$ for being explanatory. We analyze the results from the following perspectives:


\paragraph{Best Setup.} Across models, most rationales are grammatical or understandable (83\%-99\%). The best performance is achieved by Rationale BART-L, in which 80\% of the rationales were considered relevant, over 55\% correct, and between 33\% (weakeners) to 47\% (strengtheners) explanatory. Also, in general, better rationales are generated for strengthener than weakener.

\paragraph{LM and Objective.} The multi-task setup did not improve the rationale generation performance. Among the post-hoc rationalization category, Bart-based models substantially outperformed GPT2-based models. 

\paragraph{Post-hoc vs. Joint.} In the post-hoc rationalization setups, access to the target values (update more than update type) yielded more explanatory rationales  (Expl. score in Table~\ref{tab:results}), but as discussed in Section \ref{sec:analysis:generated_rationale_quality}, they are often trivial. The joint setup proved to be extremely challenging, with only 0.5\%-8.5\% of the rationales considered explanatory.

\section{Analysis}
\label{sec:analysis}
\begin{table}[t]
\scriptsize
    \centering
    \setlength\tabcolsep{2pt}
    \begin{tabular}{lllll}
    \toprule
    \textbf{Source} & \textbf{Gram.} & \textbf{Rele.} & \textbf{Corr.} & \textbf{Expl.} \\ 
    \midrule
    \textbf{Vanilla LM} & 6.54/73.03 & 4.67/14.61 & 3.74/4.49 & 0.93/2.25  \\
    \textbf{Knowledge-Enhanced LM} & 86.01/70.27 & 35.42/17.57 & 32.14/22.97 & 2.08/2.70 \\
    \textbf{COMeT} & 91.46/87.16 & 73.17/47.54 & 47.56/12.84 & 14.63/8.20 \\
    \textbf{NLI-derived} & 97.56/100.0 & 97.56/91.11 & 70.73/44.44 & 63.41/64.44 \\
    \textbf{NLI-derived w/ \/ highlights} & 50.00/84.62 & 38.24/53.85 & 17.65/30.77 & 17.65/30.77 \\
    \midrule
    \textbf{Overall} & 71.33/83.83 & 39.50/42.50 & 31.00/16.00 & 8.67/11.83 \\
    \bottomrule
    \end{tabular}
\setlength\tabcolsep{2pt}

    \vspace{-5pt}
    \caption{Human evaluation for the distant supervision rationales in the test set. Results (percents) are presented for strengtheners and weakeners separately (S/W).}
    \label{tab:extracted_rationale_human_eval}
\end{table}

\begin{table}[t]
\tiny
    \centering

\setlength\tabcolsep{2pt}
    \begin{tabular}{ll|ll}
    \toprule
    \multicolumn{2}{c|}{\textbf{Strengtheners}} & \multicolumn{2}{c}{\textbf{Weakeners}} \\
    \midrule
    \textbf{Pattern} & \textbf{\%} & \textbf{Pattern} & \textbf{\%} \\
    \midrule
    {[}S] ([H]) implies (that) [H] ({[}S]) &	64.9 & Something cannot be {[}W] and {[}H] at the same time & 33.3 \\
    {[}S] ([H]) is a rephrasing of [H] ({[}S])	& 14.9 & Something cannot be {[}W] ({[}H]) if it is {[}H] ({[}W]) & 31.8 \\
    {[}H] ({[}S]) because {[}S] ([H])	& 12.8 & {[}W] is not the same as {[}H] & 13.6 \\
    {[}S] means [H]	& 2.1 & Something is either {[}W] or {[}H] & 10.6 \\ 
    {[}S] is [H]	& 1.1 & {[}W] is not {[}H] & 6.1 \\ 
    {[}S] is the same as [H]	& 1.1 & Other & 4.6 \\ 
    Other & 3.19 \\
    \bottomrule
    \end{tabular}
\setlength\tabcolsep{6pt}
    \vspace{-5pt}
    \caption{Patterns of rationales generated by Rationale Bart-L that were considered explanatory. H, S, and W stand for Hypothesis, Strengthener and Weakener.}
    \label{tab:generated_rationale_explanatory_templates}
\end{table}

\subsection{Quality of the Distant Supervision}
\label{sec:analysis:extracted_rationale_quality}


We study the quality of rationales in the \dataset{} dataset through human evaluation. We repeated the same crowd-sourcing setup, this time evaluating the distantly supervised rationales (i.e. after filtering) of 100 random instances. 

Table~\ref{tab:extracted_rationale_human_eval} shows that the quality of the training data is surprisingly worse than that of the generated rationales. Specifically, rationales originating from LMs are often judged as incorrect and non-explanatory, much due to statements such as ``The definition of s is s''. Conversely, NLI-derived rationales are identified as the most explanatory ones, in agreement with our filtering step which kept the highest percents of NLI-derived rationales (58\%). As we show in Section \ref{sec:analysis:generated_rationale_quality}, most generated rationales are in the format of e-SNLI rationales, which might explain the discrepancy between the quality of the generated rationales and that of the training data (in which only 7.1\% of the rationales are NLI-derived rationales).

\begin{table}[t]
    \scriptsize
    \centering

\setlength\tabcolsep{2pt}
\begin{tabular}{p{0.3cm}p{8cm}}
\toprule
\multirow{3}{*}{\parbox{0.3cm}{\textbf{(1)}}} & $\mathcal{P}$: Four individuals are sitting on a small dock by the water as a boat sails by.  \\
& $\mathcal{H}$: Four people sitting near the ocean.   \\
& \textbf{W}: They're in Egypt. \\ 
& $\mathcal{R}$: Before, four people needed to go to the beach. \\
\midrule
\multirow{3}{*}{\parbox{0.3cm}{\textbf{(2)}}} & $\mathcal{P}$: Two men in orange uniforms stand before a train and do some work. \\
& $\mathcal{H}$: Tall humans working.  \\
& \textbf{S}: The men can easily touch the top of the train with their hands.  \\ 
& $\mathcal{R}$: The men can [...] train with their hands implies that they are working. \\
\midrule
\multirow{3}{*}{\parbox{0.3cm}{\textbf{(3)}}} & $\mathcal{P}$: A cyclist dressed in black and white is pointing. \\
& $\mathcal{H}$: A cyclist dressed in black and white points towards the sky. \\
& \textbf{W}: A man asked the cyclist which building is the bank. \\
& $\mathcal{R}$: Before, a cyclist needed to go to the store. \\
\midrule
\multirow{3}{*}{\parbox{0.3cm}{\textbf{(4)}}} & $\mathcal{P}$: A group of people are sight seeing and taking pictures. \\
& $\mathcal{H}$: the group is on vacation.  \\
& \textbf{S}: They are at a resort. \\ 
& $\mathcal{R}$: The group is on vacation is a rephrasing of resort. \\




\bottomrule
\end{tabular}
\setlength\tabcolsep{4pt}
    \vspace{-5pt}
    \caption{Examples for the common error types.}
    \label{tab:errore_examples}
\end{table}


\begin{table}[t]
\scriptsize
    \centering
    \begin{tabular}{lccccccc}
\toprule
\textbf{Error Type} & \textbf{(1)} & \textbf{(2)} & \textbf{(3)} & \textbf{(4)} & \textbf{(5)} & \textbf{(6)} & \textbf{(7)}\\ 
\midrule

\textbf{Strengthener} & \cellcolor{red!2.0} 4 & \cellcolor{red!10.0} 20 & \cellcolor{red!0.0} 0 & \cellcolor{red!8.0} 16 & \cellcolor{red!11.0} 22 & \cellcolor{red!9.0} 18 & \cellcolor{red!3.0} 6 \\
\textbf{Weakener} & \cellcolor{red!22.0} 44 & \cellcolor{red!7.0} 14 & \cellcolor{red!14.0} 28 & \cellcolor{red!4.0} 8 & \cellcolor{red!0.0} 0 & \cellcolor{red!0.0} 0 & \cellcolor{red!2.0} 4 \\
\textbf{Overall} & \cellcolor{red!12.0} 24 & \cellcolor{red!8.5} 17 & \cellcolor{red!7.0} 14 & \cellcolor{red!6.0} 12 & \cellcolor{red!5.5} 11 & \cellcolor{red!4.5} 9 & \cellcolor{red!2.5} 5 \\

\bottomrule
\end{tabular}
    \vspace{-5pt}
    \caption{Percent of rationales with each error type.}
    \label{tab:generated_rationale_nonexplanatory_templates}
\end{table}

\subsection{Quality of Generated Rationales}
\label{sec:analysis:generated_rationale_quality}

We manually analyzed the rationales generated by the best model (Rationale Bart-L) that were considered grammatical, relevant, and correct by humans.  

\paragraph{Explanatory.} We analyzed the 160 rationales that were considered ``explanatory'' (94 for strengtheners and 66 for weakeners), and found that almost all of them fit into one of several patterns of rationales that are trivial to generate given the target value (update type). These patterns are displayed in Table~\ref{tab:generated_rationale_explanatory_templates}. We see this as further motivation to focus on the joint setup in future research. 

\paragraph{Non-Explanatory.} We sampled and analyzed 100 rationales that were annotated as ``non-explanatory'' by workers (50-50 for strengtheners and weakeners). We found the following common types of errors, and categorized each rationale into one or more categories. The result is shown in Table~\ref{tab:generated_rationale_nonexplanatory_templates}, and exemplified in Table~\ref{tab:errore_examples}. 

\begin{enumerate}[(1)]
    \item \textbf{Insufficient}: providing one of several required reasoning hops. 
    \item \textbf{Incorrect implications}: following one of the templates in Table~\ref{tab:generated_rationale_explanatory_templates}, but not making sense.
    \item \textbf{Incorrect post/pre-conditions}: involving wrong inferences about the post-conditions of \upd{} or the pre-conditions of \hyp{}. 
    \item \textbf{Partially correct}: following a pattern in Table~\ref{tab:generated_rationale_explanatory_templates}, incorrectly using part of \upd{} or \hyp{}. For example in Table ~\ref{tab:errore_examples}, ``the group is on vacation is a rephrasing of resort'', instead of rephrasing of ``they are at a resort''. 
    \item \textbf{Repetitive statements}: defining terms or relationships between a pair of terms, by repeating the term (``The definition of $s$ is $s$'').
    \item \textbf{Wrong template}: following wrong templates in Table~\ref{tab:generated_rationale_explanatory_templates}, e.g. generating ``X is a rephrasing of Y'' when X implies Y (``The people are eating fresh seafood is a rephrasing of sitting near the ocean'').
    \item \textbf{Rationalizing the premise}: the rationale explains the premise instead of the hypothesis (e.g. ``\upd{} implies \prem{}'').
\end{enumerate}

We observe a large portion of errors (especially for weakener) are from error type (1) where the rationale needs to be completed by another hop of reasoning.

\begin{table}[t]
\scriptsize
    \centering
    \setlength\tabcolsep{6.5pt}
    \begin{tabular}{lllll}
    \toprule
    \textbf{Model} & \textbf{Gram.} & \textbf{Rele.} & \textbf{Corr.} & \textbf{Expl.} \\ 
    \midrule
    \textbf{Rationale BART-L} & 95/99.5 & 80/80.5 & 55/58 & 47/33 \\ 
    \textbf{w/o Filtering} & 99/100 & 94/93.5 & 39.5/25 & 49/26  \\
    \textbf{NLI-derived only} & 99.5/97 & 100/99 & 52.5/32.5 & 50.5/29 \\
    \bottomrule
    \end{tabular}
\setlength\tabcolsep{6pt}
    \vspace{-5pt}
    \caption{Ablation studies human evaluation. Results (percents) are presented for strengtheners and weakeners (S/W).}
    \label{tab:ablation_test_result}
\end{table}

\subsection{Ablation Studies}
\label{sec:results:ablation}

We conduct ablation studies in which we ablate either (i) the filtering step (randomly selecting a rationale from each source), or (ii) all sources besides NLI-derived rationales from our \dataset{} dataset. In both cases, we trained the best setup (Rationale Bart-L) and evaluated the results using the same human evaluation setup described in Section \ref{sec:results:human}.  

The results are reported in Table \ref{tab:ablation_test_result}. Both ablations increase the relevance of rationales while hurting their factual correctness and producing less explanatory weakener rationales. In the case of the second ablation, this is likely due to the fact that most model-generated rationales in the format of the NLI-derived rationales copy parts of the input into label-specific templates, yielding relevant but not necessarily correct or explanatory rationales.

\section{Conclusion}
\label{sec:conclusion}


We presented an approach for generating rationales for the defeasible inference task, i.e., explaining why a given update either strengthened or weakened the hypothesis. We experimented with various training setups categorized into post-hoc rationalization and joint prediction and rationalization. Rather than collecting human explanations, we chose to train our models in a distant supervision approach that requires no additional annotation cost and may generalize better across datasets. The results indicated that the post-hoc rationalization setup is easier than the joint setup, with many of the post-hoc generated rationales considered by humans as explanatory. Nonetheless, the model's success may be attributed to its access to the update type, which enabled learning a trivial mapping from the update type to rationale templates associated with it in the training data. The joint setup, on the other hand, proved to be more challenging. We hope that future work will focus on jointly predicting a label and generating a rationale, which is a more realistic setup and which may yield less trivial and more faithful rationales.

\section*{Acknowledgments}

We thank the anonymous reviewers for their insightful comments. We also thank Ana Marasovi{\'c}, Sarah Wiegreffe, xlab and Mosaic team members for helpful discussions.

\bibliography{references}

\end{document}